\pdfoutput=1
\documentclass[conference]{IEEEtran}

\usepackage{cite}
\usepackage{csquotes}
\usepackage{float}
\usepackage{amsmath,amssymb,amsfonts}
\usepackage{algorithmic}
\usepackage{graphicx}
\usepackage{textcomp}
\usepackage[svgnames, table, pdftex, dvipsnames]{xcolor}
\usepackage{xspace}
\usepackage{hyperref}
\usepackage{flushend}
\usepackage{tikz}
\usetikzlibrary{svg.path}

\newcolumntype{x}[1]{>{\centering\let\newline\\\arraybackslash\hspace{0pt}}p{#1}}
\newcolumntype{y}[1]{>{\raggedright\let\newline\\\arraybackslash\hspace{0pt}}p{#1}}
\newcolumntype{z}[1]{>{\centering\arraybackslash}m{#1}}

\newcommand\copyrighttext{%
  \footnotesize \textcopyright~2023 IEEE. Personal use of this material is permitted.
  Permission from IEEE must be obtained for all other uses, in any current or future
  media, including reprinting/republishing this material for advertising or promotional
  purposes, creating new collective works, for resale or redistribution to servers or
  lists, or reuse of any copyrighted component of this work in other works.
  }
\newcommand\copyrightnotice{%
\begin{tikzpicture}[remember picture,overlay]
\node[anchor=south,yshift=20pt] at (current page.south) {\fbox{\parbox{\dimexpr\textwidth-\fboxsep-\fboxrule\relax}{\copyrighttext}}};
\end{tikzpicture}%
}

\begin{document}
\title{A New Perspective on Evaluation Methods for Explainable Artificial Intelligence (XAI)}

\author{\IEEEauthorblockN{
Timo Speith\IEEEauthorrefmark{1}\IEEEauthorrefmark{2},
Markus Langer\IEEEauthorrefmark{3}}
\IEEEauthorblockA{\IEEEauthorrefmark{1}University of Bayreuth, Department of Philosophy, Bayreuth, Germany}
\IEEEauthorblockA{\IEEEauthorrefmark{2}Saarland University, Center for Perspicuous Computing, Saarbr\"{u}cken, Germany}
\IEEEauthorblockA{\IEEEauthorrefmark{3}Philipps-University of Marburg, Department of Psychology and Digitalization, Marburg, Germany}

Email: timo.speith@uni-bayreuth.de, markus.langer@uni-marburg.de
}

\maketitle

\copyrightnotice
\vspace{-2ex}

\begin{abstract}
One of the big challenges in the field of explainable artificial intelligence (XAI) is how to evaluate explainability approaches. Many evaluation methods (EMs) have been proposed, but a gold standard has yet to be established. Several authors classified EMs for explainability approaches into categories along aspects of the \emph{EMs themselves} (e.g., heuristic-based, human-centered, application-grounded, functionally-grounded). In this vision paper, we propose that EMs can also be classified according to aspects of the \emph{XAI process} they target. Building on models that spell out the main processes in XAI, we propose that there are \emph{explanatory information} EMs, \emph{understanding} EMs, and \emph{desiderata} EMs. This novel perspective is intended to augment the perspective of other authors by focusing less on the EMs themselves but on what explainability approaches intend to achieve (i.e., provide good explanatory information, facilitate understanding, satisfy societal desiderata). We hope that the combination of the two perspectives will allow us to more comprehensively evaluate the advantages and disadvantages of explainability approaches, helping us to make a more informed decision about which approaches to use or how to improve them.
\end{abstract}
\vspace{-1.8ex}
\begin{IEEEkeywords}
Explainability, Explainable Artificial Intelligence, XAI, Evaluation, Evaluation Methods, Metrics, Studies 
\end{IEEEkeywords}

\vspace{-0.6ex}
\section{Introduction}
With the rise of artificial intelligence (AI) in society, research on explainable AI (XAI) has experienced an upsurge in recent years \cite{BarredoArrieta2020Explainable, Langer2021What}. Explainability promises to alleviate a system's lack of transparency \cite{Chazette2020Explainability} and to provide a fruitful way to address ethical concerns about modern AI systems \cite{Langer2021Auditing, Baum2018Towards}.

Unfortunately, \enquote{explainability} is a nebulous and elusive concept that is hard to target. This causes difficulties for requirements engineers, especially when it comes to evaluating approaches to achieve explainability \cite{Langer2021What, Vilone2021Notions}. While there are several methods for evaluating explainability approaches, each with its own advantages and disadvantages, there is no consensus on what constitutes a good method \cite{Vilone2021Notions, Zhou2021Evaluating, Brunotte2022Explainability}.

In this article, we aim to address the problem by proposing a new classification approach for evaluation methods (EMs). Until now, most authors have suggested classifying EMs based on criteria specific to the EMs \emph{themselves}. However, this might ignore why people ask for explainability (e.g., to satisfy societal desiderata such as fairness and responsibility). Thus, we propose that it can also make sense to classify EMs based on the step in the \emph{XAI process} they target. Drawing on models that describe how explainability approaches attempt to satisfy societal desiderata, we suggest the existence of \emph{explanatory information}, \emph{understanding}, and \emph{desiderata} EMs.

We argue that this new perspective on EMs allows us to more comprehensively assess the individual advantages and disadvantages of EMs. Overall, we hope that by combining our classification with existing classifications in the field, we can make progress in the evaluation of explainability.

The structure of this article is as follows. In \autoref{sec:classify}, we will examine how EMs are currently classified. Based on models that describe the main processes in XAI, we will then propose and illuminate a new classification method in \autoref{sec:newperspective}. Finally, in \autoref{sec:discussion}, we will discuss our proposal and outline how it, along with existing classifications, can help advance future research on EMs.

\section{Classifying Evaluation Methods}
\label{sec:classify}

To guide the selection of appropriate EMs, their classification can help to identify the advantages and disadvantages of specific EMs. In this regard, the tripartite classification of Doshi-Velez and Kim \cite{DoshiVelez2017Towards} is widely used in XAI (see, e.g., \cite{Linardatos2020Explainable, BarredoArrieta2020Explainable, Adadi2018Peeking, Vilone2021Notions}). They distinguish between \emph{application-grounded}, \emph{human-grounded}, and \emph{functionally-grounded} EMs. In more detail, these three types of EMs are:

\begin{itemize}
    \item \textbf{Application-Grounded} EMs evaluate how explainability approaches affect human \emph{experts} in \emph{specific} tasks or applications.
    \item \textbf{Human-Grounded} EMs evaluate how explainability approaches affect \emph{an arbitrary} human in a \emph{general} setting.
    \item \textbf{Functionally-Grounded} EMs use  mathematical specifications and tests to evaluate explainability approaches. They do not involve humans or empirical studies.
\end{itemize}

This classification is based on the properties of the EMs \emph{themselves}. For instance, application-grounded and human-grounded EMs involve user studies to evaluate explainability approaches, whereas functionally-grounded EMs focus on proving mathematical properties of explainability approaches. 

Other authors have suggested dyadic classifications. Bibal and Fr\'{e}nay, for example, propose to distinguish between \emph{heuristic-based} and \emph{user-based} EMs \cite{Bibal2016Interpretability}. For them, EMs from Doshi-Velez and Kim's categories application-grounded and human-grounded would be in the user-based category, while functionally-grounded EMs would correspond to heuristic-based EMs. The case is analogous for the categories \emph{human-centered} and \emph{objective}, which Vilone and Longo propose \cite{Vilone2021Notions}.

In addition to proposing these categorizations, most authors discuss the respective advantages and disadvantages of EMs in their proposed categories. For example, application-grounded EMs are time-consuming and costly, but provide insights in the actual effects that explainability approaches have in the real world and with human experts who use AI-based systems. 

Furthermore, human-grounded evaluations, while involving a less specific participant sample (and thus providing little insights for specialized application environments), are a cost-effective means of gaining insights into how humans process information generated by explainability approaches. 

Finally, functionally-grounded EMs are probably the least costly. Because they rely on formal specifications, they can be conducted without participants, thus avoiding common challenges with studies involving humans (e.g., low number of participants, biased samples, replicability and generalizability issues, limitations to study designs). However, the absence of human participants is not only their strength, but also their weakness: since many explainability approaches are ultimately intended to help human actors in some way, evaluations without participants may ultimately fail to assess these goals.

\section{A New Perspective on Evaluation Methods}
\label{sec:newperspective}

Although the categorizations discussed above already provide vital insights regarding the advantages and disadvantages of individual EMs, they only focus on categorizing EMs according to properties of EMs themselves. However, it is widely acknowledged that explainability approaches eventually aim to satisfy \emph{societal desiderata}\footnote{We follow Langer et al. and use the term \enquote{desiderata} to collectively refer to stakeholders' interests, goals, expectations, needs, and demands regarding AI systems \cite{Langer2021What}. Other authors have used the term \enquote{quality aspects} \cite{Chazette2021Exploring}.} (e.g., assign responsibility for a decision \cite{Heinrichs2020Evidence, Baum2022Responsibility} or assess its fairness \cite{Langer2021Auditing}) \cite{Langer2021What, BarredoArrieta2020Explainable, Chazette2021Exploring}.

In other words, a classification based on the EMs themselves may neglect why explainability approaches were used in the first place. For instance, if we use EMs from all the categories provided by Doshi-Velez and Kim, we may have a broad picture of whether human experts or laypersons find (certain aspects of) an explainability approach useful, or whether it satisfies certain mathematical properties. However, this does not mean that we know whether this explanability approach satisfies relevant social desiderata (or why, exactly, it failed).

The required insights may be gained by evaluating the individual steps by which explainability approaches aim to satisfy societal desiderata. For this reason, we propose a new perspective on EMs that is inspired by the models of Hofmann et al. \cite{Hoffman2018Metrics} and Langer et al. \cite{Langer2021What}, which attempt to capture the key processes and concepts between explainability approaches and the satisfaction of societal desiderata.

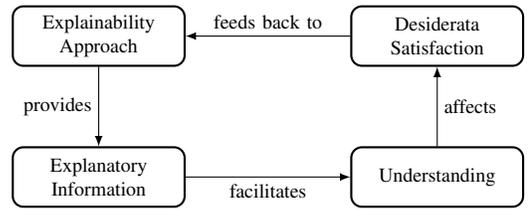
\begin{figure}[htbp]
\centering
\begin{tikzpicture}[scale=0.75, transform shape,
    block1/.style={rectangle, draw=black, thick, text width=8em, align=center, rounded corners, minimum height=3em},
    block2/.style={rectangle, draw=black, thick, text width=6em, align=center, rounded corners, minimum height=3em}]
    
    \node[draw, block1] at (-3, 0)   (xai) {Explainability Approach};
    \node[draw, block1] at (-3, -2.5)    (exp) {Explanatory Information};
    \node[draw, block1] at (3, -2.5) (und) {Understanding};
    
    \node[draw, block1] at (3, 0)    (des) {Desiderata Satisfaction};
    
    \draw[-latex] (xai) -- (exp)  node[midway, left] (fac) {provides};
    \draw[-latex] (exp) -- (und)  node[midway, below] (let) {facilitates};
    \draw[-latex] (und) -- (des)  node[midway, right] (sat) {affects};
    \draw[-latex] (des) -- (xai)  node[midway, above] (fbt) {feeds back to};
    
\end{tikzpicture}
\caption{A simplified version of the explainability models that Hofmann et al. \cite{Hoffman2018Metrics} and Langer et al. \cite{Langer2021What} have proposed.}
\label{fig:model}
\end{figure}

According to these models, \emph{explainability approaches} provide \emph{explanatory information} with the aim of facilitating people's \emph{understanding}. This understanding, in turn, affects the satisfaction of \emph{societal desiderata}. Depending on the final status of desiderata satisfaction, an explainability approach may need to be adapted, or a new one chosen or even developed (see \autoref{fig:model} for a simplified version of the models).  

Since these models describe the process through which explainability approaches aim to satisfy stakeholders' desiderata, they can be used to derive implications for the evaluation of such approaches. Specifically, it might be fruitful to use the concepts from these models to guide the evaluation of explainability approaches. In line with this thought, our perspective on EMs of explainability approaches is based on which component of the models an EM addresses.

Hence, we distinguish between methods that focus on ex\-pla\-na\-to\-ry information (\emph{explanatory information} EMs), methods that focus on understanding (\emph{understanding} EMs), and methods that focus on desiderata satisfaction (\emph{desiderata} EMs).

Below, we will briefly describe each of these categories with their individual advantages and disadvantages. While doing so, we will enrich them with examples sorted in the previous classifications (we follow Vilone and Longo \cite{Vilone2021Notions} and use the categories \enquote{human-centered} and \enquote{objective})\footnote{To do so, we combine the categories \enquote{application-grounded} \cite{DoshiVelez2017Towards}, \enquote{human-groun\-ded} \cite{DoshiVelez2017Towards}, \enquote{user-based} \cite{Bibal2016Interpretability}, and \enquote{human-centered} \cite{Vilone2021Notions} on the one hand, and \enquote{functionally-grounded} \cite{DoshiVelez2017Towards}, \enquote{heuristic-based} \cite{Bibal2016Interpretability},  and \enquote{obejctive} \cite{Vilone2021Notions} on the other. We do not want to imply that human-centered EMs do not strive for objectivity, but rather that they rely on some form of human feedback.}. This serves to show that our classification is orthogonal to previous ones.

\subsection{Explanatory Information Evaluation Methods}

The rationale behind \emph{explanatory information} EMs is that if explanatory information provided by explainability approaches accurately describe system-related aspects\footnote{In the XAI debate, there are many possibilities for the exact aspect that is supposed to be understood. For our argument, however, it is only important that the understanding is concerned with some aspect of a system.} or if recipients perceive the explanatory information to be useful and comprehensible, it should successfully facilitate understanding.

Regarding \emph{objective} criteria to judge the quality of explanatory information, one of the most crucial ones is \emph{fidelity} \cite{Rosenfeld2019Explainability, Zhou2021Evaluating, Velmurugan2021Developing, Speith2022Evaluate, Vilone2021Notions}. Fidelity means that the explanatory information provided offers insights regarding \emph{actual} decision processes in systems. Note that a high fidelity does not mean that \emph{all} factors contributing to a decisions are mentioned. This is covered by another criterion, namely \emph{completeness} \cite{Gevrey2003Review}.

The \emph{explanation satisfaction scale} by Hoffman et al. \cite{Hoffman2018Metrics} is an example of a \emph{human-centered} explanatory information EM. In this scale, participants respond to items on a scale between 1 (\enquote{I disagree strongly}) and 5 (\enquote{I agree strongly}) assessing the quality of explanatory information.\footnote{Partly, the scale by Hoffman et al. \cite{Hoffman2018Metrics} is also an understanding EM because it includes items that test for system-related understanding.} Sample items are \enquote{This explanation of how the [software, algorithm, tool] works is satisfying} and \enquote{This explanation of how the [software, algorithm, tool] works has sufficient detail.} Responding \enquote{I agree strongly} for these items would indicate that participants are satisfied with the explanatory information provided.

Explanatory information EMs acknowledge that it is vital to assess whether explainability approaches produce information that is comprehensible and of high quality. This is crucial in evaluating explainability approaches. For example, if an approach provides information that is hard to grasp or unsatisfying, it will likely fail in facilitating understanding of any aspects related to a system.

Nevertheless, exclusively evaluating explanatory information does not conclusively reveal whether users better understand system-related aspects. Specifically, we cannot deduce a user's understanding of system-related aspects from the fidelity of explanatory information or from their own judgments about whether they are satisfied with this information \cite{Rozenblit2002Misunderstood}.

Overall, explanatory information EMs target an early step in the models proposed by Hoffman et al. or Langer et al. \cite{Langer2021What, Hoffman2018Metrics} without directly assessing whether explanatory information facilitated the next step (i.e., understanding).

\subsection{Understanding Evaluation Methods}

The rationale behind \emph{understanding} EMs is that a central goal of explainability approaches is to facilitate understanding of system-related aspects \cite{Paez2019Pragmatic, Koehl2019Explainability}. Consequently, these methods examine whether explanatory information helped people better understand such aspects. As with explanatory information EMs, there are both objective and human-centered means of assessing understanding.

One well-known \emph{objective} understanding EM is the \emph{size of the model} \cite{Gacto2011Interpretability, Freitas2014Comprehensible, Garcia2009Study, Otero2016Improving}. The rough idea is that the larger a model is (e.g., the more parameters it has), the more difficult it is to understand. Accordingly, if the aim is to explain an opaque model with an understandable surrogate model, or if the aim is to produce an inherently understandable model, it is assumed that such models ought to be small \cite{Belle2021Principles}.

Regarding \emph{human-centered} understanding EMs, one approach is to ask participants whether, after receiving information, they better understood system-related aspects (e.g., \enquote{From the explanation, I understand how the [software, algorithm, tool] works.}, \cite{Hoffman2018Metrics}). Note that this is different from evaluating explanatory information (and, thereby, from explanatory information EMs), where participants evaluate the information itself, rather than the understanding it induced. 

Another understanding EM would be to ask people questions about things they may have learned about the system (e.g., which factors most strongly affect its outputs) and to examine whether they correctly responded to these question.\footnote{Note that such an evaluation may be considered \enquote{objective} in the purely psychological literature as it tries to assess understanding based on objective criteria. In our classification we include this as a human-centered evaluation method because it still builds on tasks that human participants perform.}

Understanding EMs acknowledge that it is crucial to assess whether explainability approaches have facilitated a person's understanding. In line with this, the aforementioned methods provide efficient ways to approximate whether there will be any effects regarding understanding. However, relying on the size of the model or on participants to report their own understanding has limitations. Concerning the size of the model, there are no agreed upon standards as to what is acceptable. Furthermore, the size has been proven not to reliably indicate the model's understandability (e.g., some large models can be more comprehensible than smaller ones \cite{Freitas2014Comprehensible}).

When it comes to participants reporting their understanding of system-related aspects, it is debatable whether people are willing to state or able to adequately assess whether they understood certain system-related aspects better after receiving explanatory information \cite{Keil2006Explanation}. 
In other words, it may be premature to conclude that an explainability approach has successfully facilitated understanding when participants report on a scale from \enquote{I disagree strongly} to \enquote{I agree strongly} that they \enquote{agree strongly} that explanatory information has helped them better understand a system's internal processes. 
Similarly, participants might sometimes have an unjustified belief that the information increased their understanding of system-related aspects. In these cases, follow-up questions may find that these participants provide incorrect answers; they only had an illusion of understanding \cite{Rozenblit2002Misunderstood, Trout2007Psychology}.

In sum, the understanding EMs above provide only partly reliable information regarding participants' actual understanding of system-related aspects.

\subsection{Desiderata Evaluation Methods}

The rationale behind \emph{desiderata} EMs is that we can assess the success of explainability approaches by directly measuring important outcome variables (i.e., desiderata) such as trust, acceptance, and human-machine performance \cite{Hoffman2018Metrics, Mueller2019Explanation, Mueller2021Principles}. The logic is that explainability approaches are successful when they affect the satisfaction of such desiderata \cite{Hoffman2018Metrics, Langer2021What}. As far as we know, there are only human-centered desiderata EMs.

Given that there are various desiderata, there is also a plethora of desiderata EMs. Depending on the desideratum in question, the EM differs. For instance, trust is often evaluated via self-report items, while human-machine performance is often evaluated via work efficiency tests \cite{Hoffman2018Metrics}.\footnote{For more information, see Mueller et al. \cite{Mueller2019Explanation, Mueller2021Principles}, who provide overviews on evaluation methods for various desiderata (e.g., trust, satisfaction). See also \autoref{tab:classification} for an overview of some studies evaluating different desiderata.}

Evaluating desiderata satisfaction is valuable because it directly assesses whether the goals that people have when using explainability approaches are achieved. However, desiderata EMs may reach their limits in revealing \emph{why} an explainability approach has affected a desideratum or \emph{why} it has failed to to do so. These limitations emerge because desiderata EMs implicitly assume that successful explainability approaches lead to an increased understanding of system-related aspects which affects these outcome variables \cite{Mueller2021Principles}. This assumption, however, is rarely tested. 

\renewcommand{\arraystretch}{1.2}
\rowcolors{2}{blue!20}{white}
\begin{table*}[htbp]
    \centering
    \begin{tabular}{|z{0.11\textwidth}|z{0.4\textwidth}|z{0.4\textwidth}|}
        \hline
        \rowcolor{blue!50}
        Type of EM & Objective & Human-Centered  \\
        \hline
        Explanatory Information & 
        \begin{itemize}
            \item (in)fidelity: \cite{Yeh2019Fidelity, Rosenfeld2019Explainability, Zhou2021Evaluating, Velmurugan2021Developing, Speith2022Evaluate, Vilone2021Notions}
            \item sensitivity to input perturbation: \cite{Nguyen2020Model, Arras2016Explaining, Binder2016Analyzing, Samek2017Explainable, Samek2017Evaluating, Arras2017Relevant}; e.g., robustness: \cite{Ghorbani2019Interpretation, Alvarez-Melis2018Robustness, Kindermans2019Reliability}
            \item sensitivity to model parameter randomisation: \cite{Adebayo2018Local, Adebayo2018Sanity}; e.g., Sensitivity-n: \cite{Ancona2018Towards}
            \item explanation completeness: \cite{Gevrey2003Review}
            \item text quality: \cite{Barratt2017Interpnet} (e.g., BLEU, METEOR, CIDEr)
            \vspace{-2.4ex}
        \end{itemize} &
        \begin{itemize}
            \item questionnaires: \cite{Lim2009Why, Lim2009Assessing}; e.g., explanation satisfaction scale \cite{Hoffman2018Metrics}
            \item comparison tests: \cite{Holzinger2019Kandinsky}
            \vspace{-2.4ex}
        \end{itemize}
        \\
        Understanding & 
        \begin{itemize}
            \item size of the model: \cite{Gacto2011Interpretability, Freitas2014Comprehensible, Garcia2009Study, Otero2016Improving}
            \item filter interpretability: \cite{Zhang2018Visual}
            \vspace{-2.4ex}
        \end{itemize} &
        \begin{itemize}
            \item questionnaires: \cite{Kaur2020Interpreting}; e.g., explanation satisfaction scale \cite{Hoffman2018Metrics}
            \item user tasks: \cite{Piltaver2014Comprehensibility} e.g., assessment tasks: \cite{Stock2018Convnets, Allahyari2011User, Piltaver2014Comprehensibility, Lage2018Human}; prediction tasks: \cite{Poursabzi2021Manipulating, Huysmans2011Empirical, Piltaver2014Comprehensibility, Ribeiro2018Anchors}
            \item interviews: \cite{Tullio2007Works}
            \vspace{-2.4ex}
        \end{itemize}\\
        Desiderata & (none identified) &
        \begin{itemize}
            \item confidence: \cite{Suermondt1993Evaluation, Ye1995Impact}
            \item debugging: \cite{Vellido2012Making, Srinivasan2017Interpretable, Lapuschkin2016Analyzing, Kulesza2015Principles, Kulesza2011Why-Oriented, Krause2016Interacting}
            \item human-machine performance: e.g., work-efficiency tests: \cite{Hoffman2018Metrics}
            \item trust: \cite{Dzindolet2003Role}; e.g., self-reports: \cite{Hoffman2018Metrics}
            \item usability: e.g., system causability scale: \cite{Holzinger2020Measuring}
            \item usefulness: \cite{Krause2016Interacting}
            \vspace{-2.4ex}
        \end{itemize}\\
        \hline
    \end{tabular}
    \medskip
    \caption{Papers that use or propose EMs, classified according to a combination of previous proposals and our proposal.}
    \label{tab:classification}
\end{table*}

Instead, desiderata EMs directly assess the satisfaction of desiderata without considering a person's understanding of system-related aspects. For instance, it could be that incorporating an explainability approach into a system increases trust not by augmenting a person's understanding, but by making it appear as if there is a justification for an output \cite{Eiband2019Impact}. Trust that is formed in this way may, in fact, be unwarranted
\cite{Jacovi2021Formalizing}. In such cases, the EM could come to misleading results.

In sum, while desiderata EMs may prove sufficient in some cases (e.g., when desiderata are well-specified and satisfied), they are limited in providing feedback in cases of failure.

\section{Discussion and Future Work}
\label{sec:discussion}

The above examples are by no means representative of all EMs in XAI. Still, they illustrate that our proposal augments existing classifications. As we have shown by the objective/human-centered distinction, classifications suggested by other authors are orthogonal to our proposed classification. 

Our new classification proposal corroborates the fact that each EM has its advantages and disadvantages. Thus, in order to comprehensively assess the quality of explainability approaches, the different aspects that are important for EMs need to be combined. Whereas the classification of previous authors implies that it makes sense to consider human-centered and objective EMs, our proposal implies that we need to assess explanatory information, understanding, as well as desiderata satisfaction to assess the quality of an explainability approach.

With this idea in mind, we want to take a first step towards painting a more complete picture. To this end, we used the review conducted by Vilone and Longo \cite{Vilone2021Notions} to categorize further EMs in the six categories emerging from combining previous classification proposals and our proposal (see Table \ref{tab:classification}). In this way, we were able to find studies or proposals for five of the six categories, with the exception being objective EMs for desiderata. Overall, it is noticeable that objective EMs tend to concentrate on the explanations themselves, while human-centered EMs focus on understanding and desiderata.

If an extensive literature analysis were to confirm our initial findings that there are no or only a few objective EMs for desiderata, our classification would have made it possible to identify gaps in research. Similar thoughts apply to human-centered EMs for explanatory information. To conduct exactly such a further analysis is a goal for further research.

Even if it turns out that neither of these two findings can be confirmed in a more comprehensive analysis, employing a combination of diverse classifications can still provide a more comprehensive understanding of the landscape of EMs in XAI. In particular, combining the two perspectives on classification offers synergies because they show different advantages and disadvantages of individual EMs: one perspective can tell what the methods themselves can and cannot do, the other can say what aspects of the XAI process the method evaluates.

Future research needs to demonstrate how, precisely, the insights outlined here on the advantages and disadvantages of different EMs, as they become apparent through the different classifications, can be utilized to determine which EM should be used in a specific context. Only then can it be justified which explainability approach is the appropriate choice for that context. Perhaps this will even lead to a gold standard.

\section{Conclusion}

Through our new perspective on EMs, we hope to advance the discussion regarding the evaluation of XAI. Even if there will never be a gold standard, because the best evaluation procedure varies by context and consists of a combination of several EMs, our classification may allow us to better identify the appropriate combination of methods for a given context. 

\section*{Acknowledgments}
Work on this paper was funded by the Volkswagen Foundation grants AZ 98509, 98513, and 98514 \href{https://explainable-intelligent.systems}{\enquote{Explainable Intelligent Systems}} (EIS) and by the DFG grant 389792660 as part of \href{https://perspicuous-computing.science}{TRR~248}, especially project A6. We thank three anonymous reviewers for their helpful feedback.

\bibliographystyle{IEEEtran}
\bibliography{bibliography}

\end{document}